\documentclass[runningheads]{llncs}
\usepackage{graphicx}
\usepackage{comment}
\usepackage{amsmath,amssymb} 
\usepackage{color}
\usepackage{hyperref}
\usepackage{booktabs}
\usepackage[width=122mm,left=12mm,paperwidth=146mm,height=193mm,top=12mm,paperheight=217mm]{geometry}
\makeatletter
\newcommand{\printfnsymbol}[1]{%
	\textsuperscript{\@fnsymbol{#1}}%
}
\makeatother

\begin{document}
\pagestyle{empty}
\mainmatter
\def\ECCVSubNumber{1449}  

\title{TuiGAN: Learning Versatile Image-to-Image Translation with Two Unpaired Images} 

\titlerunning{xx} 
\authorrunning{xx} 
\author{ Jianxin Lin\textsuperscript{\rm 1}\thanks{The first two authors contributed equally to this work}, 
	Yingxue Pang\textsuperscript{\rm 1}\printfnsymbol{1}, Yingce Xia\textsuperscript{\rm 2}, 
	Zhibo Chen\textsuperscript{\rm 1}, Jiebo Luo\textsuperscript{\rm 3}}
\institute{
	$^1$University of Science and Technology of China \\
	$^2$Microsoft Research Asia  \quad	$^3$University of Rochester\\
\email{\{linjx, pangyx\}@mail.ustc.edu.cn \quad
	yingce.xia@microsoft.com \\
	chenzhibo@ustc.edu.cn \quad jluo@cs.rochester.edu} }

\maketitle

\begin{abstract}
An unsupervised image-to-image translation (UI2I) task deals with learning a mapping between two domains without paired images. While existing UI2I methods usually require numerous unpaired images from different domains for training, there are many scenarios where training data is quite limited. In this paper, we argue that even if each domain contains a single image, UI2I can still be achieved. To this end, we propose TuiGAN, a generative model that is trained on only two unpaired images and amounts to one-shot unsupervised learning.  With TuiGAN, an image is translated in a coarse-to-fine manner where the generated image is gradually refined from global structures to local details. We conduct extensive experiments to verify that our versatile method can outperform strong baselines on a wide variety of UI2I tasks. Moreover, TuiGAN is capable of achieving comparable performance with the state-of-the-art UI2I models trained with sufficient data. Our code is available at \url{https://github.com/linjx-ustc1106/TuiGAN-PyTorch}.

		
\end{abstract}

\section{Introduction}
Unsupervised image-to-image translation (UI2I) tasks aim to map images from a source domain to a target domain with the main source content preserved and the target style transferred,  while no paired data is available to train the models. Recent UI2I methods have achieved remarkable successes  \cite{liu2017unsupervised,lee2018diverse,zhu2017unpaired,lin2018conditional,choi2019stargan}. Among them, conditional UI2I gets much attention, where two images are given: an image from the source domain used to provide the main content, and the other one from the target domain used to specify which style the main content should be converted to. To achieve UI2I, typically one needs to collect numerous unpaired images from both the source and target domains. 
	 
However, we often come across cases for which there might not be enough unpaired data to train the image translator. An extreme case resembles one-shot unsupervised learning, where only one image in the source domain and one image in the target domain are given but unpaired.  Such a scenario has a wide range of real-world applications, e.g.,  taking a photo and then converting it to a specific style of a given picture, or replacing objects in an image with target objects for image manipulation. In this paper, we take the first step towards this direction and study UI2I given only two unpaired images.
	
\begin{figure}[!t]
\centering
\includegraphics[width=11cm]{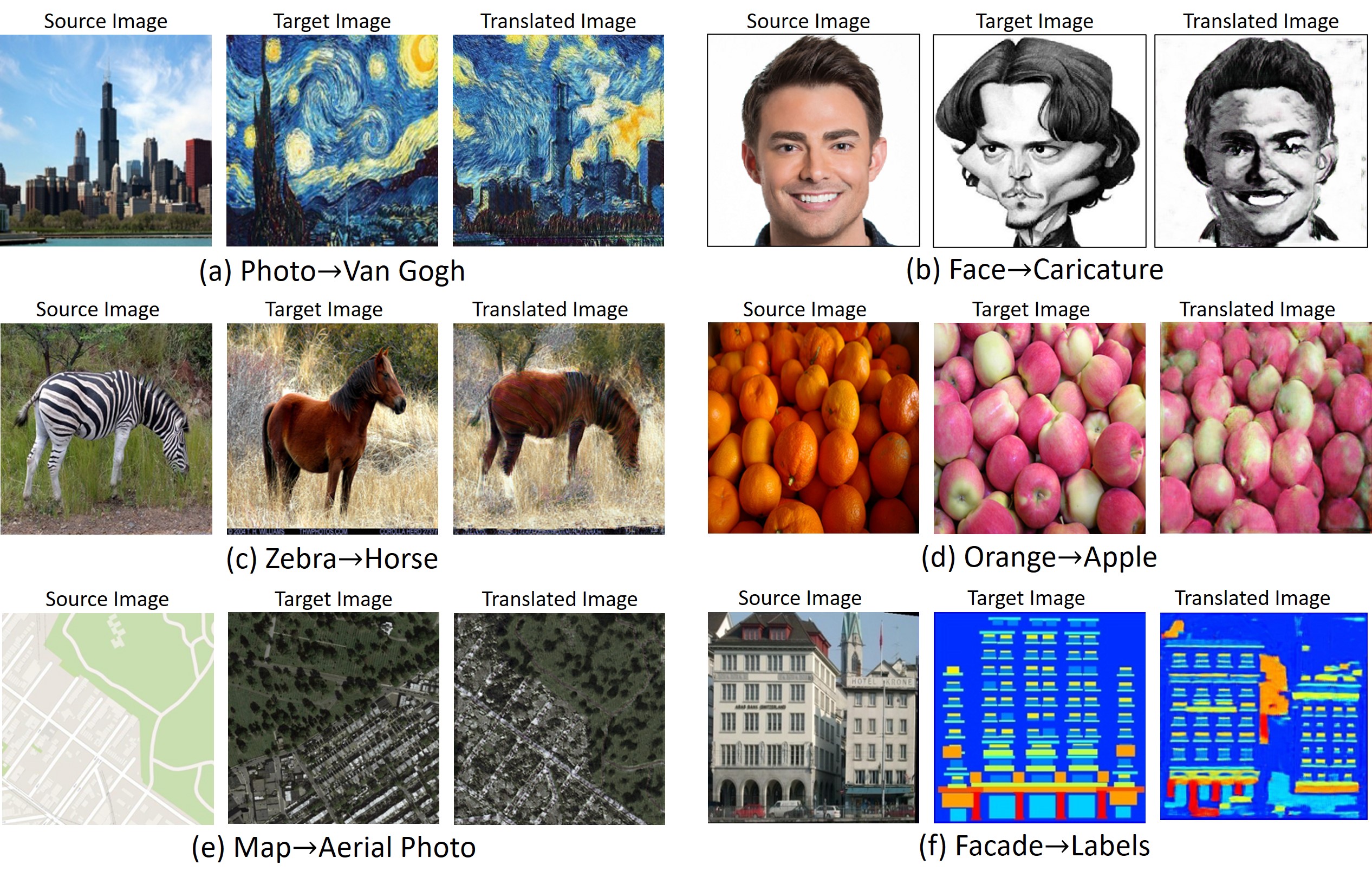}
\caption{Several results of our proposed method on various tasks ranging from image style transfer (Figures (a), (b)) to object transformation (Figures (c), (d)) and appearance transformation (Figures (e), (f)). In each sub-figure, the three pictures from left to right refer to the source image (providing the main content), target image (providing the style and high-level semantic information), and translated image.}
\label{fig:examples}
\end{figure}

	
Note that the above problem subsumes the conventional image style transfer task. Both problems require one source image and one target image, which serve as the content and style images, respectively. In image style transfer, the features used to describe the styles (such as the Gram matrix of pre-trained deep features~\cite{gatys2016image}) of the translated image and the style image should match (e.g., Fig.~\ref{fig:examples}(a)). In our generalized problem, not only the style but the higher-level semantic information should also match. As shown in Fig. \ref{fig:examples}(c), on the zebra-to-horse translation, not only the background style (e.g., prairie) is transferred, but the high-level semantics (i.e., the profile of the zebra) is also changed. 
	
	
Achieving UI2I requires the models to effectively capture the variations of domain distributions between two domains, which is the biggest challenge for our problem since there are only two images available. To realize such one-shot translation, we propose a new conditional generative adversarial network, TuiGAN, which is able to transfer the domain distribution of input image to the target domain by progressively translating image from coarse to fine.  The progressive translation enables the model to extract the underlying relationship between two images by continuously varying the receptive fields at different scales. Specifically, we use two pyramids of generators and discriminators to refines the generated result progressively from global structures to local details. For each pair of generators at the same scale, they are responsible for producing images that look like the target domain ones. For each pair of discriminators at the same scale, they are responsible for capturing the domain distributions of the two domains at the current scale. The ``one-shot'' term in our paper is different from the ones in  \cite{benaim2018one,cohen2019bidirectional}, which use a single image from  the source domain and a set of images from the target domain for UI2I. In contrast, we only use two unpaired images from two domains in our work. 

We conduct extensive experimental validation with comparisons to various baseline approaches using various UI2I tasks, including horse $\leftrightarrow$ zebra, facade $\leftrightarrow$ labels, aerial maps $\leftrightarrow$ maps,  apple $\leftrightarrow$ orange, and so on. The experimental results show that the versatile approach effectively addresses the problem of one-shot image translation. We show that our model can not only outperform existing UI2I models in the one-shot scenario, but more remarkably, also achieve comparable performance with UI2I models trained with sufficient data.

Our contributions can be summarized as follows:
\begin{itemize}
\item We propose a TuiGAN to realize image-to-image translation with only two unpaired images.
\item We leverage two pyramids of conditional GANs to progressively translate image from coarse to fine.
\item We demonstrate that the a wide range of UI2I tasks can be tackled using our versatile model.
\end{itemize}
\section{Related Works}
\subsection{Image-to-Image Translation}


The earliest concept of image-to-image translation (I2I) may be raised in \cite{hertzmann2001image} which supports a wide variety of ``image filter'' effects. Rosales et al. \cite{rosales2003unsupervised} propose to infer correspondences between a source image and another target image using Bayesian framework. With the development of deep neural networks, the advent of Generative Adversarial Networks (GAN) \cite{goodfellow2014generative} really inspires many works in I2I.  Isola et al. \cite{isola2017image} propose a conditional GAN called  ``pix2pix'' model for a wide range of supervised I2I tasks. However, paired data may be difficult or even impossible to obtain in many cases. 
DiscoGAN  \cite{kim2017learning}, CycleGAN \cite{zhu2017unpaired} and DualGAN \cite{yi2017dualgan} are proposed to tackle the unsupervised image-to-image translation (UI2I) problem by constraining two cross-domain translation models to maintain cycle-consistency. Liu et al. \cite{liu2019few} propose a FUNIT model for few-shot UI2I. However, FUNIT requires not only a large amount of training data and computation resources to infer unseen domains, but also the training data and unseen domains to share similar attributes. Our work does not require any pre-training and specific form of data. Related to our work, Benaim et al. \cite{benaim2018one} and Cohen et al. \cite{cohen2019bidirectional} propose to solve the one-shot cross-domain translation problem, which aims to learn an unidirectional mapping function given a single image from the source domain and a set of images from the target domain. Moreover, their methods cannot translate images in the opposite direction as they claim that one seen sample in the target domain is difficult for capturing domain distribution. However, in this work, we focus on solving UI2I given only two unpaired image from two domains and realizing I2I in both directions. Concurrently to our work, Benaim et al. \cite{benaim2020structural} also attempts to learn transformation between two unpaired images by using multi-scale generative models.

\subsection{Image Style Transfer}
Image style transfer can be traced back to Hertzmann et al.'s work \cite{hertzmann1998painterly}. More recent approaches use neural networks to learn the style statistics. Gatys et al.  \cite{gatys2016image} first model image style transfer by minimizing the Gram matrix of pre-trained deep features. Luan et al. \cite{luan2017deep} further propose to realize photorealistic style transfer which can preserve the photorealism of the content image. To avoid inconsistent stylizations in semantically uniform regions, Li et al. \cite{li2018closed} introduce a two-step framework in which both steps have a closed-form solution.  However, it is difficult for these models to transfer higher-level semantic structures, such as object transformation. We demonstrate that our model can outperform Li et al. \cite{li2018closed} in various UI2I tasks.

\subsection{Single Image Generative Models}
Single image generative models aim to capture the internal distribution of an image. Conditional GAN based models have been proposed for texture expansion \cite{zhou2018non} and image retargeting \cite{shocher2018ingan}. InGAN \cite{shocher2018ingan}  is trained with a single natural input and learns its internal patch-distribution by an image-specific GAN.  Unconditional GAN based models also have been proposed for texture synthesis \cite{bergmann2017learning,li2016precomputed,jetchev2016texture} and image manipulation \cite{shaham2019singan}. In particular, SinGAN \cite{shaham2019singan} employs an unconditional pyramidal generative model to learn the patch distribution based on images of different scales. However, these single image generative models usually take one image into consideration and do not capture the relationship between two images. In contrast, our model aims to capture the distribution variations between two unpaired images. In this way, our model can transfer an image from a source distribution to a target distribution while maintaining its internal content consistency. 

\begin{figure}[t!]
	\centering
	\includegraphics[width=10cm]{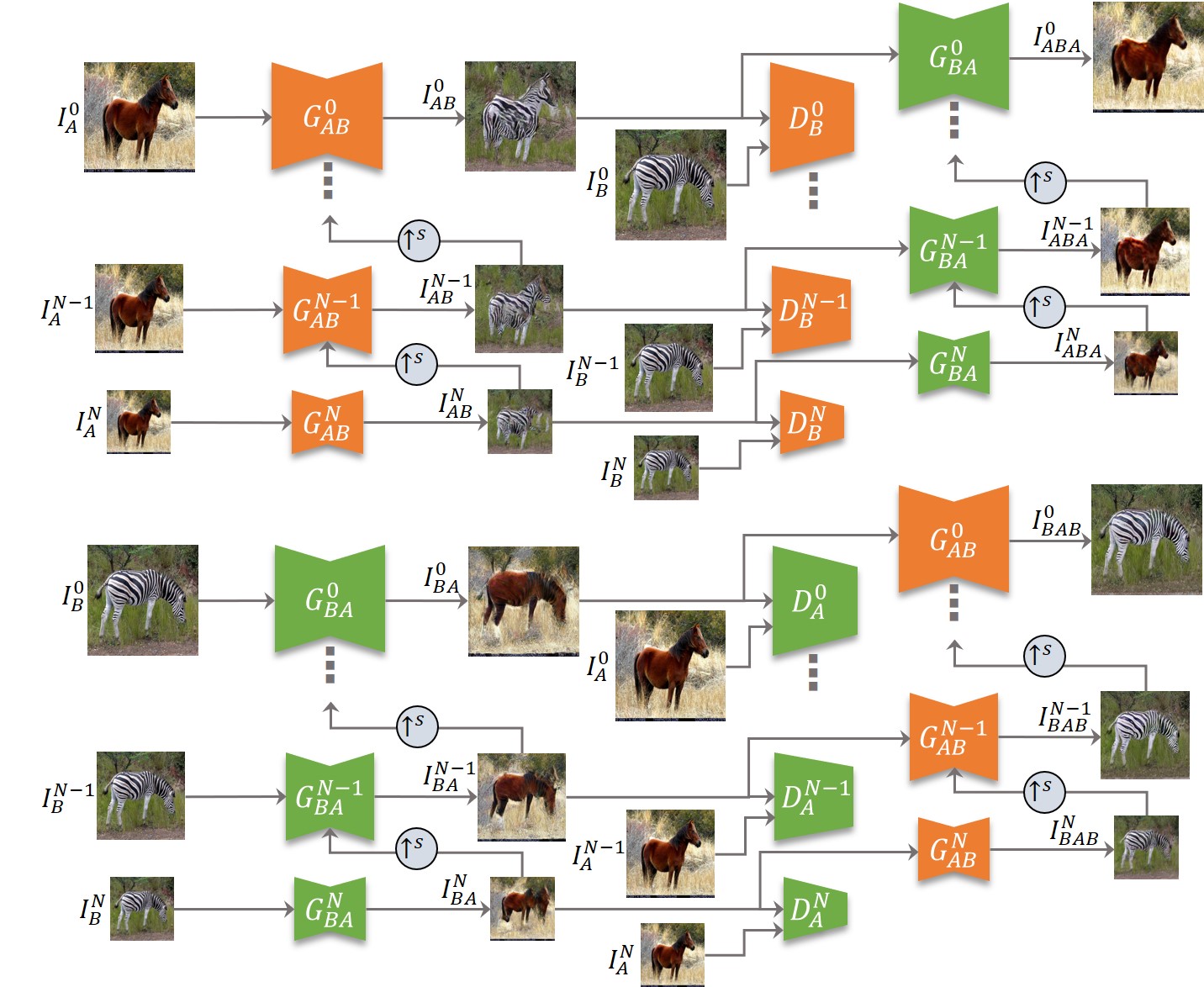}
	\caption{\textbf{TuiGAN network architecture:}  TuiGAN consists of two symmetric pyramids of generators ($G^{n}_{AB}$s and $G^{n}_{BA}$s) and discriminators ($D^{n}_{B}$s and $D^{n}_{A}$s), $0\leq n \leq N$. At each scale, the generators take the downsampled source image  and previously translated image to generate the new translated image.  The discriminators learn the domain distribution by progressively narrowing the receptive fields. The whole framework is learned in a scale-to-scale fashion and the final result is obtained at the finest scale. }
	\label{fig:framework}
\end{figure}

\section{Method}
Given two images $I_{A}\in{A}$ and $I_{B}\in{B}$, where $A$ and $B$ are two image domains, our goal is to convert $I_{A}$ to $I_{AB}\in{B}$ and $I_{B}$ to $I_{BA}\in{A}$ without any other data accessible. Since we have only two unpaired images, the translated result (e.g., $I_{AB}$) should inherit the domain-invariant features of the source image (e.g., $I_{A}$) and replace the domain-specific features with the ones of the target image (e.g., $I_{B}$) \cite{zhu2017unpaired,lee2018diverse,huang2018multimodal}. To realize such image translation, we need to obtain a pair of mapping functions $G_{AB}:A\mapsto B$ and $G_{BA}:B\mapsto A$, such that
\begin{equation}
I_{AB}=G_{AB}(I_{A}), \quad I_{BA}=G_{BA}(I_{B}).
\end{equation}
Our formulation aims to learn the internal domain distribution variation between $I_{A}$ and $I_{B}$. Considering that the training data is quite limited, $G_{AB}$ and $G_{BA}$ are implemented as two multi-scale conditional GANs that progressively translate images from coarse to fine. In this way, the training data can be fully leveraged at different resolution scales. We downsample $I_A$ and $I_B$ to $N$ different scales, and then obtain $\mathcal{I}_A=\{I_A^n|n=0,1,\cdots,N\}$ and $\mathcal{I}_B=\{I_B^n|n=0,1,\cdots,N\}$, where $I_A^n$ and $I_B^n$ are downsampled from $I_A$ and $I_B$, respectively, by a scale factor $(1/s)^n$ ($s\in\mathbb{R}$). 

In previous literature, multi-scale architectures have been explored for unconditional image generation with multiple training images~\cite{karras2017progressive,karras2019style,denton2015deep,huang2017stacked}, conditional image generation with multiple paired training images~\cite{wang2018high} and image generation with a single training image~\cite{shaham2019singan}. In this paper, we leverage the benefit of multi-scale architecture for one-shot unsupervised learning, in which only two unpaired images are used to learn UI2I.

\subsection{Network Architecture}
The network architecture of the proposed TuiGAN is shown in Fig. \ref{fig:framework}. The entire framework consists of two symmetric translation models: $G_{AB}$ for $I_{A}\to{I_{AB}}$ (the top part in Fig.\ref{fig:framework}) and $G_{BA}$ for $I_{B}\to{I_{BA}}$ (the bottom part in Fig. \ref{fig:framework}).  $G_{AB}$ and $G_{BA}$ are made up of a series of generators, $\{G_{AB}^n\}_{n=0}^N$ and $\{G_{BA}^n\}_{n=0}^{N}$, which can achieve image translation at the corresponding scales. At each image scale, we also need  discriminators $D_A^n$ and $D_B^n$ ($n\in\{0,1,\cdots,N\}$), which is used to verify whether the input image is a natural one in the corresponding domain.




\noindent{\bf Progressive Translation}
The translation starts from images with the lowest resolution and gradually moves to the higher resolutions. $G^N_{AB}$ and $G^N_{BA}$ first map $I_A^N$ and $I_B^N$ to the corresponding target domains: 
\begin{equation}
I^{N}_{AB}=G^{N}_{AB}(I^{N}_{A});\;I^{N}_{BA}=G^N_{BA}(I^N_B).
\end{equation}
For images with scales $n<N$, the generator $G^{n}_{AB}$ has two inputs, $I^{n}_{A}$ and the previously generated $I^{n+1}_{AB}$. Similarly, $G^{n}_{BA}$ takes $I^{n}_{B}$ and $I^{n+1}_{BA}$ as inputs. Mathematically,
\begin{equation}
I^{n}_{AB}=G^{n}_{AB}(I^{n}_{A},I^{n+1\uparrow}_{AB}),\;I^{n}_{BA}=G^{n}_{BA}(I^{n}_{B},I^{n+1\uparrow}_{BA}),
\label{eq:generator_n}
\end{equation}
where $\uparrow$ means to use bicubic upsampling to resize image by a scale factor $s$. Leveraging $I^{n+1}_{AB}$, $G^{n}_{AB}$ could refine the previous output with more details, and $I^{n+1}_{AB}$ also provides the global structure of the target image for current resolution. Eqn.\eqref{eq:generator_n} is iteratively applied until the eventual output $I^0_{AB}$ and $I^0_{BA}$ are obtained.




\begin{figure}[!t]
\centering
\includegraphics[width=12cm]{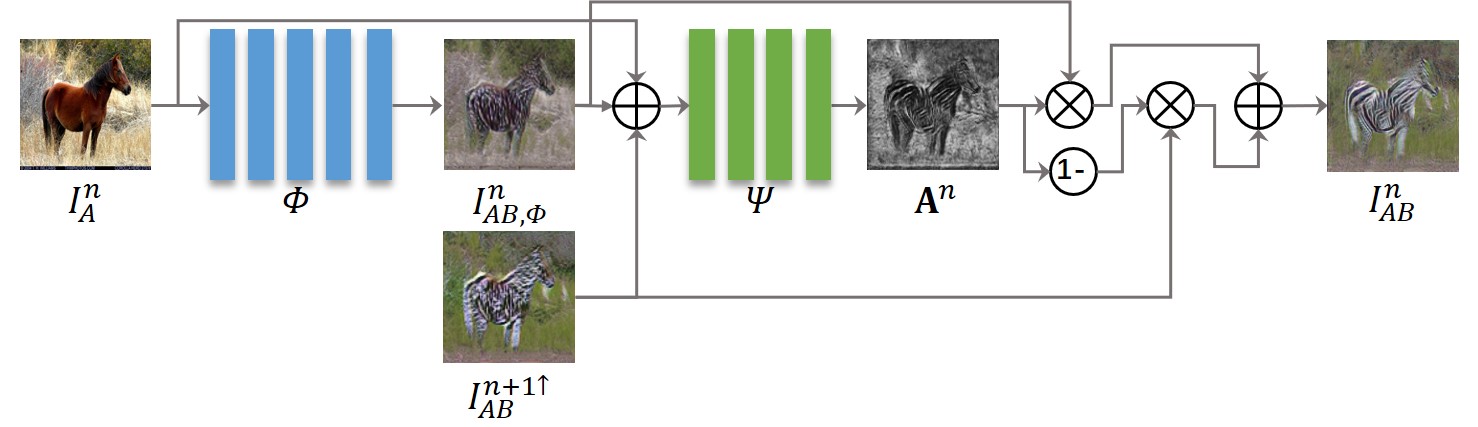}
\caption{Architecture of the generator $G^{n}_{AB}$, which achieves the $I^n_{A}\to{I^n_{AB}}$ translation. There are two modules, $\Phi$ and $\Psi$. The input $I^n_A$ is first transformerd via $\Phi$ to obtain $I^n_{AB,\Phi}$. Then, the transformed $I^n_{AB,\Phi}$, original input $I^n_A$ and the output of previous scale $I^{n+1\uparrow}_{AB}$ are fused by model $\Psi$ to generated a mask $\mathbf{A}^n$. Finally, $I^n_{AB,\Phi}$ and $I^{n+1\uparrow}_{AB}$ are linearly combined through $\mathbf{A}^n$ to obtain the final output.
}
\label{fig:fusion}
\end{figure}

\noindent{\bf Scale-aware Generator}
The network architecture of $G^{n}_{AB}$ is shown in Fig.~\ref{fig:fusion}. Note that $G^{n}_{AB}$ and $G^{n}_{BA}$ shares the same architecture but have different weights. $G^{n}_{AB}$ consists of two fully convolutional networks. Mathematically, $G^n_{AB}$ works as follows:
\begin{equation}
\begin{aligned}
& I^{n}_{AB,\Phi}=\Phi({I^{n}_{A}}),\;\mathbf{A}^{n}=\Psi{(I^{n}_{AB,\Phi},I^{n}_{A},I^{n+1\uparrow}_{AB})}, \\
& I^{n}_{AB} =  \mathbf{A}^{n}\otimes I^{n}_{AB,\Phi}+(1-\mathbf{A}^{n}) \otimes I^{n+1\uparrow}_{AB},
\end{aligned}
\label{eq:gen_module}
\end{equation}
where $\otimes$ represents pixel-wise multiplication. As shown in Eqn.\eqref{eq:gen_module}, we first use $\Phi$ to preprocess $I^n_A$ into $I^n_{AB,\Phi}$ as the initial translation. Then, we use an attention model $\Psi$ to generate a mask $\mathbf{A}^n$, which models long term and multi-scale dependencies across image regions \cite{zhang2019self,pumarola2018ganimation}. $\Psi$ takes $I^{n}_{AB,\Phi}$, $I^{n+1\uparrow}_{AB}$ and $I^{n}_{A}$ as inputs and outputs $\mathbf{A}^n$ considering to balance two scales' results. Finally, $I^n_{AB,\Phi}$ and $I^{n+1\uparrow}_{AB}$ are linearly combined through the generated $\mathbf{A}^n$ to get the output $I^n_{AB}$.

Similarly, the translation $I_{B}\to{I_{BA}}$ at $n$-th scale is implemented as follows:
\begin{equation}
\begin{aligned}
&I^{n}_{BA,\Phi}=\Phi({I^{n}_{B}});\;\mathbf{A}^{n}=\Psi{(I^{n}_{BA,\Phi},I^{n}_{B},I^{n+1\uparrow}_{BA})}, \\
&I^{n}_{BA} =  \mathbf{A}^{n}\otimes I^{n}_{BA,\Phi}+(1-\mathbf{A}^{n}) \otimes I^{n+1\uparrow}_{BA}.
\end{aligned}
\end{equation}
In this way, the generator focuses on regions of the image that are responsible of synthesizing details in current scale and keeps the previously learned global structure untouched in the previous scale. As shown in Fig.~\ref{fig:fusion}, the previous generator has generated global structure of a zebra in $I^{n+1\uparrow}_{AB}$, but still fails to generate stripe details. In the $n$-th scale, the current generator generates an attention map to add stripe details on the zebra and produces better result $I^{n}_{AB}$.

\subsection{Loss Functions}  

Our model is progressively trained from low resolution to high resolution. Each scale keeps fixed after training. For any $n\in\{0,1,\cdots,N\}$, the overall loss function of the $n$-th scale is defined as follows:
\begin{equation}
\mathcal{L}^{n}_{\text{ALL}}=\mathcal{L}^{n}_{\text{ADV}}+\lambda_{\text{CYC}}\mathcal{L}^{n}_{\text{CYC}}+\lambda_{\text{IDT}}\mathcal{L}^{n}_{\text{IDT}}+\lambda_{\text{TV}}\mathcal{L}^{n}_{\text{TV}},
\end{equation}
where $\mathcal{L}^{n}_{\text{ADV}}$, $\mathcal{L}^{n}_{\text{CYC}}$, $\mathcal{L}^{n}_{\text{IDT}}$, $\mathcal{L}^{n}_{\text{TV}}$ refer to adversarial loss, cycle-consistency loss, identity loss and total variation loss respectively, and $\lambda_{\text{CYC}}$, $\lambda_{\text{IDT}}$, $\lambda_{\text{TV}}$ are  hyper-parameters to balance the tradeoff among each loss term. At each scale, the generators aim to minimize $\mathcal{L}^{n}_{\text{ALL}}$ while the discriminators is trained to maximize $\mathcal{L}^{n}_{\text{ALL}}$. We will introduce details of these loss functions.

\noindent{\bf Adversarial Loss}   
The adversarial loss builds upon that fact that the discriminator tries to distinguish real images from synthetic images and generator tries to fool the discriminator by generating realistic images. At each scale $n$, there are two discriminators $D_A^n$ and $D_B^n$, which take an image as input and output the probability that the input is a natural image in the corresponding domain. We choose WGAN-GP \cite{gulrajani2017improved} as adversarial loss  which can effectively improve the stability of adversarial training by weight clipping and gradient penalty:
\begin{equation}
\begin{aligned}
\mathcal{L}^n_{\text{ADV}}  &= D^n_{B}(I^n_B)-D^n_{B}(G^n_{AB}(I^n_A)) + D^n_{A}(I^n_A)-D^n_{A}(G^n_{BA}(I^n_B)) \\
&-\lambda_{\text{PEN}} (\Vert \nabla_{\hat{I}^n_B}D^n_{B}(\hat{I}^n_B)  \Vert_2-1)^2-\lambda_{\text{PEN}} (\Vert \nabla_{\hat{I}^n_A}D^n_{A}(\hat{I}^n_A)  \Vert_2-1)^2,
\end{aligned}
\end{equation}
where $\hat{I}^n_B=\alpha I^n_B+(1-\alpha) I^n_{AB}$ , $\hat{I}^n_A=\alpha I^n_A+(1-\alpha) I^n_{BA}$ with $\alpha \sim U(0,1)$, $\lambda_{\text{PEN}}$ is the penalty coefficient.

\noindent{\bf Cycle-Consistency Loss} One of the training problems of conditional GAN is mode collapse, i.e., a generator produces an especially plausible output whatever the input is.   We utilize cycle-consistency loss \cite{zhu2017unpaired} to constrain the model to retain the inherent properties of input image after translation: $\forall n\in\{0,1,\cdots,N\}$,
\begin{equation}
\begin{aligned}
& \mathcal{L}^{n}_{\text{CYC}}=\|{I^{n}_{A}-I^{n}_{ABA}}\|_{1}+\|{I^{n}_{B}-I^{n}_{BAB}}\|_{1},\quad\text{where}\\
& I^{n}_{ABA}=G^{n}_{BA}(I^{n}_{AB},I^{n+1\uparrow}_{ABA}),\; I^{n}_{BAB}=G^{n}_{AB}(I^{n}_{BA},I^{n+1\uparrow}_{BAB}),&\text{ if }n<N;\\
& I^{N}_{ABA}=G^{N}_{BA}(I^{N}_{AB}), \; I^{N}_{BAB}=G^{N}_{AB}(I^{N}_{BA}),&\text{ if }n=N.
\end{aligned}
\end{equation}


\noindent{\bf Identity Loss}  
We noticed that relying on the two losses mentioned above for one-shot image translation  could easily lead to color \cite{zhu2017unpaired} and texture misaligned results.  To tackle the problem, we introduce the identity loss at each scale, which is denoted as $L^{n}_{\text{IDT}}$. Mathematically,
\begin{equation}
\begin{aligned}
&\mathcal{L}^{n}_{\text{IDT}}=\Vert{I^{n}_{A}-I^{n}_{AA}}\Vert_{1}+\Vert{I^{n}_{B}-I^{n}_{BB}}\Vert_{1},\quad\text{where}\\
& I^{n}_{AA}=G^{n}_{BA}(I^{n}_{A},I^{n+1\uparrow}_{AA}),\; I^{n}_{BB}=G^{n}_{AB}(I^{n}_{B},I^{n+1\uparrow}_{BB}),&\text{if }n<N;\\
& I^{N}_{AA}=G^{N}_{BA}(I^{N}_{A}), \quad I^{N}_{BB}=G^{N}_{AB}(I^{N}_{B}),&\text{if }n=N.
\end{aligned}
\end{equation}
We found that identity loss can effectively preserve the consistency of color and texture tone between the input and the output images as shown in Section \ref{sec:ablation}.

\noindent{\bf Total Variation Loss}
To avoid noisy and overly pixelated, following~\cite{mahendran2015understanding}, we introduce total variation (TV) loss to help in removing rough texture of the generated image and get more spatial continuous and smoother result.  It encourages images to consist of several patches by calculating the differences of neighboring pixel values in the image. Let $x[i,j]$ denote the pixel located in the $i$-th row and $j$-th column of image $x$. The TV loss at the $n$-th scale is defined as follows:
\begin{align}
& \mathcal{L}^{n}_{\text{TV}}=L_{tv}(I^{n}_{AB})+L_{tv}(I^{n}_{BA}),\\
& L_{tv}(x)=\sum_{i,j}\sqrt{(x[i,j+1]-x[i,j])^{2}+(x[i+1,j]-x[i,j])^{2}},\;x\in\{I^{n}_{AB},I^{n}_{BA}\}\nonumber.
\end{align}
\subsection{Implementation Details}
\noindent{\bf Network Architecture}
As mentioned before, all generators share the same architecture and they are all fully convolutional networks. In detail, $\Phi$ is constructed by 5 blocks of the form 3x3 Conv-BatchNorm-LeakyReLU \cite{ioffe2015batch} with stride 1. $\Psi$ is constructed by 4 blocks of the form 3x3 Conv-BatchNorm-LeakyReLU. For each discriminator, we use the Markovian discriminator (PatchGANs) \cite{isola2017image} which has the same 11x11 patch-size as $\Phi$ to keep the same receptive field as generator.  

\noindent{\bf Training Settings}
We train our networks using Adam \cite{kingma2014adam} with initial learning rate $0.0005$, and we decay the learning rate after every 1600 iterations. We set our scale factor $s=4/3$ and train 4000 iterations for each scale. The number of scale $N$ is set to 4.  For all experiments, we set weight parameters $\lambda_{\text{CYC}}=1$, $\lambda_{\text{IDT}}=1$, $\lambda_{\text{TV}}=0.1$ and  $\lambda_{\text{PEN}}=0.1$. Our model requires 3-4 Hrs on a single 2080-Ti GPU with the images of 250$\times$250 size.
\section{Experiments}
We conduct experiments on several tasks of unsupervised image-to-image translation, including the general UI2I tasks\footnote{In this paper, we refer to general UI2I as tasks  where there are multiple images in the source and target domains, i.e., the translation tasks studied in~\cite{zhu2017unpaired}.}, image style transfer, animal face translation and paint-to-image translation, to verify our versatile TuiGAN. To construct datasets of one-shot image translation, given a specific task (like horse$\leftrightarrow$zebra translation~\cite{zhu2017unpaired}), we randomly sample an image from the source domain and the other one from the target domain, respectively, and train models on the selected data.

\subsection{Baselines}
We compare TuiGAN with two types of baselines. The first type leverages the full training data without subsampling. We choose CycleGAN~\cite{zhu2017unpaired} and DRIT~\cite{lee2018diverse} algorithms for image synthesis. The second type leverages partial data, even one or two images only. We choose the following baselines: 

\noindent(1) OST~\cite{benaim2018one}, where one image from the source domain and a set of images in the target domain are given; 

\noindent(2) SinGAN~\cite{shaham2019singan}, which is a pyramidal unconditional generative model trained on only one image from the target domain, and injects an image from the source domain to the trained model for image translation.

\noindent(3) PhotoWCT~\cite{li2018closed}, which can be considered as a special kind of image-to-image translation model, where a content photo is transferred to the reference photo's style while remaining photorealistic. 

\noindent(4) FUNIT~\cite{liu2019few}, which targets few-shot UI2I and requires lots of data for pre-training. We test the one-shot translation of FUNIT. 

\noindent(5) ArtStyle~\cite{gatys2015neural}, which is a classical art style transfer model.

For all the above baselines, we use their official released code to produce the results.

\subsection{Evaluation Metrics}
\noindent(1) \textbf{Single Image Fr\'{e}chet Inception Distance (SIFID)  \cite{shaham2019singan}}: SIFID captures the  difference of internal distributions between two images, which is implemented by computing the Fr\'{e}chet Inception Distance (FID) between deep features of two images. A lower SIFID score indicates that the style of two images is more similar. We compute SIFID between translated image and corresponding target image.

\noindent(2) \textbf{Perceptual Distance (PD) \cite{johnson2016perceptual}}: PD computes the perceptual distance between images.  A lower PD score indicates that the content of two images is more similar. We compute PD between translated image and corresponding source image.

\noindent(3) \textbf{User Preference (UP)}: We conduct user preference studies for performance evaluation since the qualitative assessment is highly subjective. 

\begin{figure}[t!]
\centering
\includegraphics[width=12cm]{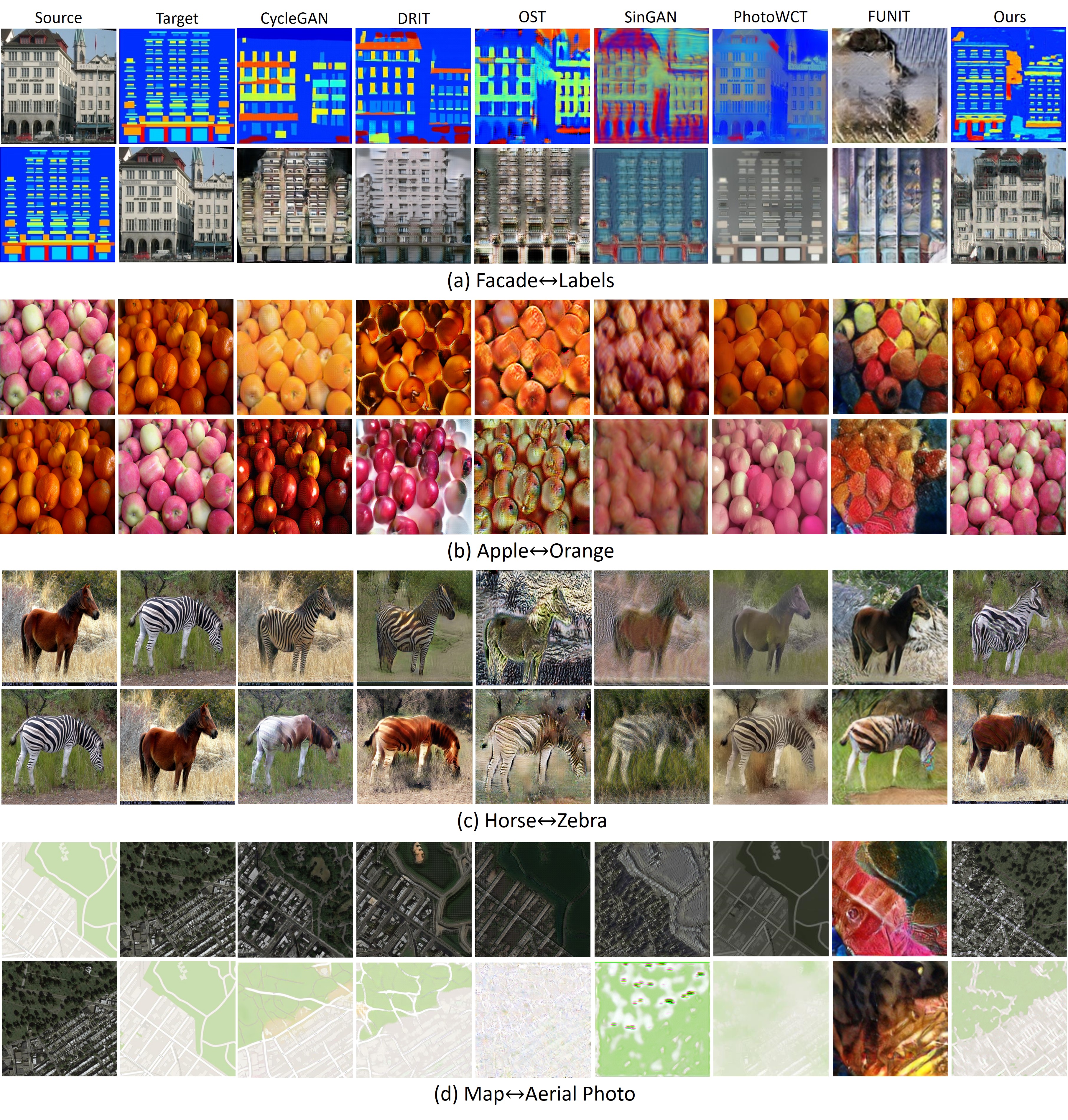}
\caption{Results of general UI2I tasks using CycleGAN (trained with full training dataset), DRIT (trained with the full training dataset), OST (trained with 1 sample in the source domain and full data in the target domain), SinGAN (trained with one target image), PhotoWCT (trained with two unpaired images), FUNIT (pre-trained) and our TuiGAN (trained with two unpaired images).}
\label{fig:comparisons}
\end{figure}

\subsection{Results}
\subsubsection{General UI2I Tasks}
Following~\cite{zhu2017unpaired}, we first conduct general experiments on Facade$\leftrightarrow$Labels, Apple$\leftrightarrow$Orange, Horse$\leftrightarrow$Zebra and Map$\leftrightarrow$Aerial Photo translation tasks to verify the effectiveness of our algorithm. The visual results of our proposed TuiGAN and the baselines are shown in Fig.~\ref{fig:comparisons}. 

Overall, the images generated by TuiGAN exhibit better translation quality than OST, SinGAN, PhotoWCT and FUNIT. While both SinGAN and PhotoWCT change global colors of the source image, they fail to transfer the high-level semantic structures as our model (e.g., in Facade$\leftrightarrow$Labels and Horse$\leftrightarrow$Zebra). Although OST is trained with the full training set of the target domain and transfers high-level semantic structures in some cases, the generated results contain many noticeable artifacts, e.g., the irregular noises on apples and oranges. Compared with CycleGAN and DRIT trained on full datasets, TuiGAN achieves comparable results to them. There are some cases that TuiGAN produces better results than these two models in Labels$\rightarrow$Facade,  Zebra$\rightarrow$Horse tasks, which further verifies that our model can actually capture domain distributions with only two unpaired images. 
	
The results of average SIFID, PD and UP are reported in Table~\ref{table:quantitative}. For user preference study, we randomly select 8 unpaired images, and generate 8 translated images for each general UI2I task. In total, we collect 32 translated images for each subject to evaluate. We display the source image, target image and two translated images from our model and another baseline method respectively on a webpage in random order. We ask each subject to select the better translated image at each page.  We finally collect the feedback from 18 subjects of total 576 votes and 96 votes for each comparison. We compute the percentage from a method is selected as the User Preference (UP) score. 

\setlength{\tabcolsep}{4pt}
\begin{table}[!t]
\centering
\caption{Average SIFID, PD and UP across different general UI2I tasks.}
\resizebox{\textwidth}{10mm}{
\begin{tabular}{cccccccc}
\toprule
Metrics & CycleGAN & DRIT & OST & SinGAN & PhotoWCT & FUNIT & Ours \\
\midrule
SIFID ($ \times 10^{-2}$) & 0.091 & 0.142 & 0.123 & 0.384 & 717.622 & 1510.494 & 0.080\\
PD & 5.56 & 8.24 & 10.26 & 7.55 & 3.27 & 7.55 & 7.28 \\
UP & 61.45\% & 52.08\% & 26.04\%& 6.25\%& 25.00\%& 2.08\%& -\\
\bottomrule
\end{tabular}
}
\label{table:quantitative}
\end{table}
\setlength{\tabcolsep}{1.4pt}

We can see that TuiGAN obtains the best SIFID score among all the baselines, which shows that our model successfully captures the distributions of images in the target domain. In addition, our model achieves the third place in PD score after CycleGAN and PhotoWCT. From the visual results, we can see that PhotoWCT can only change global colors of the source image, which is the reason why it achieves the best PD score. As for user study, we can see that most of the users prefer the translation results generated by TuiGAN than OST, SinGAN, PhotoWCT and FUNIT. Compared with DRIT trained on full data, our model also achieves similar votes from subjects.

\begin{figure}[t!]
\centering
\includegraphics[width=12cm]{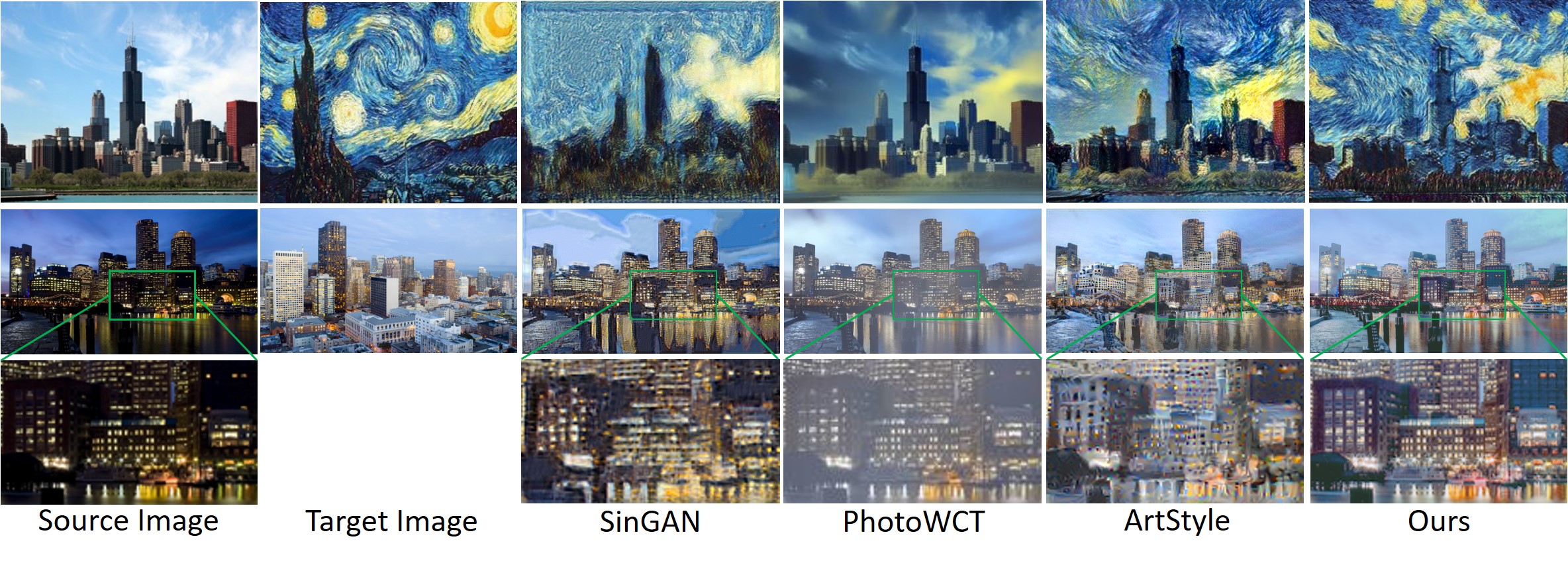}
\caption{Results of image style transfer. The first row represents the results of art style transfer, and the second row is the results of photorealistic style transfer. We amplify the green boxes in photorealistic style transfer results at the third row to show more details.   }
\label{fig:vangogh}
\end{figure}
\subsubsection{Image Style Transfer}\label{sec:image_style_trans}
We demonstrate the effectiveness of our TuiGAN on image style transfer: 
art style transfer, which is to convert image to the target artistic style with specific strokes or textures, and photorealistic style transfer, which is to obtain stylized photo that remains photorealistic. Results are shown in Fig. \ref{fig:vangogh}. As can be seen in the first row of Fig.\ref{fig:vangogh}, TuiGAN retains the architectural contour and generates stylized result with vivid strokes, which just looks like Van Gogh’s painting. Instead, SinGAN fails to generate clear stylized image, and PhotoWCT \cite{li2018closed} only changes the colors of real photo without capturing the salient painting patterns. In the second row, we transfer the night image to photorealistic day image with the key semantic information retained. Although SinGAN and ArtStyle produce realistic style, they fail to the maintain detailed edges and structures. The result of PhotoWCT is also not as clean as ours. Overall, our model achieves competitive performance on both types of image style transfer, while other methods usually can only target on a specific task but fail in another one. 
	
\begin{figure}[t!]
\centering
\includegraphics[width=12cm]{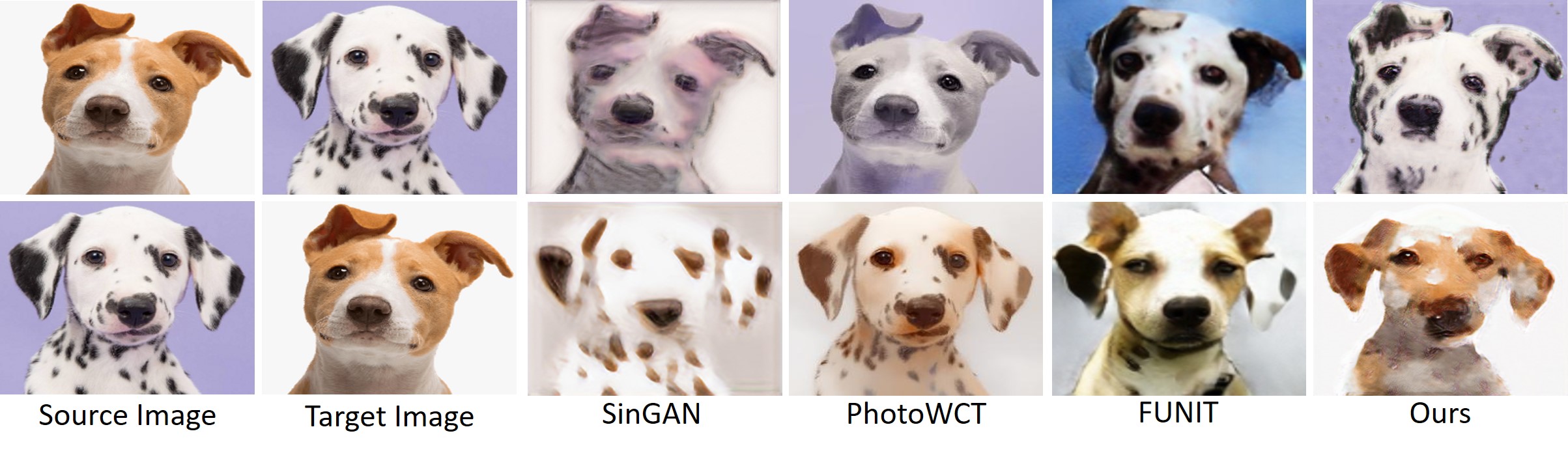}
\caption{Results of animal face translation. Our model can accurately transfer the fur colors, while FUNIT, a model pre-trained on animal face dataset, does not work as well as our model.}
\label{fig:dog}
\end{figure}

\noindent{\bf Animal Face Translation}
To compare with the few-shot model FUNIT, which is pretained on animal face dataset, we conduct the animal face translation experiments as shown in Fig.\ref{fig:dog}. We also include SinGAN and PhotoWCT for comparison. As we can see, our model can better transfer the fur colors from image in the target domain to the that of the source domain than other baselines: SinGAN \cite{shaham2019singan} generates results with faint artifacts and blurred dog shape; PhotoWCT \cite{li2018closed} can not transfer high-level style feature (e.g. spots) from the target image although it preserves the content well; and FUNIT generates results that are not consistent with the target dog's appearance.
	

\begin{figure}[t!]
\centering
\includegraphics[width=12cm]{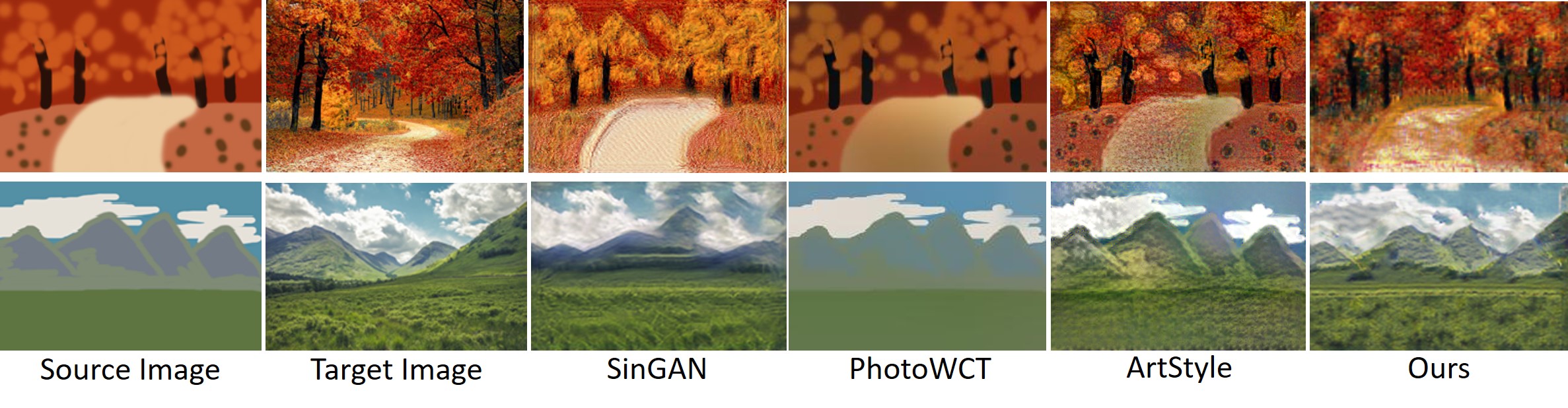}
\caption{Results of painting-to-image translation. TuiGAN can translate more specific style patterns of the target image (e.g., leaves on the road in the first row) and maintain more accurate content of the source images (e.g., mountains and clouds in the second row).
}
\label{fig:trees}
\end{figure}
\noindent{\bf Painting-to-Image Translation}\label{paint_to_image}
This task focuses to generate photo-realistic image with more details based on a roughly related clipart as described in SinGAN \cite{shaham2019singan}. We use the two samples provided by SinGAN for comparison.  The results are shown in Fig.\ref{fig:trees}. Although two testing images share similar elements (e.g., trees and road), their styles are extremely different. Therefore, PhotoWCT and ArtStyle fail to transfer the target style in two translation cases. SinGAN also fails to generate specific details, such as leaves on the road in the first row of Fig.\ref{fig:trees}, and maintain accurate content, such as mountains and clouds in the second row of Fig.\ref{fig:trees}. Instead, our method preserves the crucial components of input and generates rich local details in two cases. 
	

\begin{figure}[t!]
\centering
\includegraphics[width=12cm]{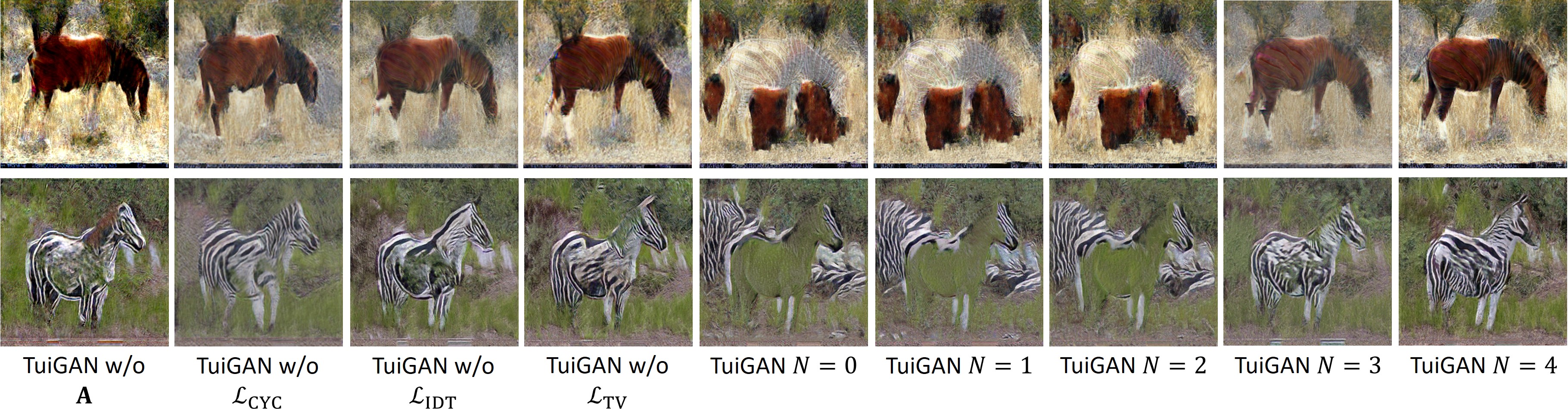}
\caption{Visual results of ablation study.}
\label{fig:ablation}
\end{figure}

\subsection{Ablation Study}\label{sec:ablation}
To investigate the influences of different training losses, generator architecture and multi-scale structure, we conduct several ablation studies based on Horse$\leftrightarrow$Zebra task. Specifically, 

\noindent(1) Fixing $N=4$, we remove the cycle-consistent loss (TuiGAN w/o $L_{\text{CYC}}$), identity loss (TuiGAN w/o $L_{\text{IDT}}$), total variation loss (TuiGAN w/o $L_{\text{TV}}$) and compare the differences.

\noindent(2) We range $N$ from $0$ to $4$ to see the effect of different scales. When $N=0$, our model can be roughly viewed as the CycleGAN \cite{zhu2017unpaired} that is trained with two unpaired images.

\noindent(3) We remove the attention model $\Psi$ in the generators, and combine $I^{n}_{AB,\Phi}$ and $I^{n+1\uparrow}_{AB}$ by simply addition (briefly denoted as TuiGAN w/o $\mathbf{A}$). 

The qualitative results are shown in Fig.\ref{fig:ablation}. Without $L_{\text{IDT}}$, the generated results suffers from inaccurate color and texture (e.g., green color on the transferred zebra). Without attention mechanism or $L_{\text{CYC}}$, our model can not guarantee the completeness of the object shape (e.g., missed legs in the transferred horse).  Without $L_{\text{TV}}$, our model produces images with artifacts (e.g., colour spots around the horse). The results from $N=0$ to $N=3$ either have poor global content information contained (e.g. the horse layout) or have obvious artifacts (e.g. the zebra stripes). Our full model (TuiGAN $N=4$) could capture the salient content of the source image and transfer remarkable style patterns of the  target image. 
	
We compute the quantitative ablations by assessing SIFID and PD scores of different variants of TuiGAN.  As shown in Table~\ref{table:ablation}, our full model still obtains the lowest SIFID score and PD score, which indicates that our TuiGAN could generate more realistic and stylized outputs while preserving the content unchanged.

\setlength{\tabcolsep}{4pt}
\begin{table}[t!]
\begin{center}
\caption{Quantitative comparisons between different variants of TuiGAN in terms of SIFID and PD scores. The best scores are in bold. }
\label{table:ablation}
\resizebox{\textwidth}{19mm}{
\begin{tabular}{cccccccccc}
\toprule
&&  &  & TuiGAN &  &  &  &  & \\
Metrics& w/o $\mathbf{A}$ & w/o $\mathcal{L}_{\text{CYC}}$ & w/o $ \mathcal{L}_{\text{IDT}}$ & w/o $\mathcal{L}_{\text{TV}}$ & $N=0$ & $N=1$ & $N=2$ & $N=3$  & $N=4$ \\
				\midrule
				\begin{tabular}[c]{@{}c@{}}SIFID ($ \times 10^{-4}$)\\ Horse$\rightarrow$Zebra\end{tabular}& 1.08 & 3.29 & 2.43 & 2.41 & 2.26 & 2.32 & 2.31 & 2.38 & \textbf{1.03}\\
				\begin{tabular}[c]{@{}c@{}}SIFID ($ \times 10^{-4}$)\\ Zebra$\rightarrow$Horse\end{tabular} & 2.09 & 5.61 & 5.54 & 10.85 & 3.75 & 3.86 & 3.77 & 6.30 & \textbf{1.79}\\
				\begin{tabular}[c]{@{}c@{}}PD\\ Horse$\rightarrow$Zebra\end{tabular} & 8.00 & 6.98 & 8.24 & 6.90 & 6.40 & 6.82 & 6.76 & 6.25 & \textbf{6.16}\\
				\begin{tabular}[c]{@{}c@{}}PD\\ Zebra$\rightarrow$Horse\end{tabular} & 10.77 & 7.92 & 8.00 &  6.48 & 7.77 & 7.92 & 8.68 & 6.87 & \textbf{5.91}\\
				\bottomrule
			\end{tabular}}
		\end{center}
	\end{table}
	\setlength{\tabcolsep}{1.4pt}

\section{Conclusion} 
In this paper, we propose TuiGAN, a versatile conditional generative model that is trained on only two unpaired image, for image-to-image translation. Our model is designed in a coarse-to-fine manner, in which two pyramids of conditional GANs refine the result progressively from global structures to local details. In addition, a scale-aware generator is introduced to better combine two scales' results. We validate the capability of TuiGAN on a wide variety of unsupervised image-to-image translation tasks by comparing with several strong baselines. Ablation studies also demonstrate that  the losses and network scales are reasonably designed. Our work represents a further step toward the possibility of unsupervised learning with extremely limited data.

	\clearpage
	%
	%
\bibliographystyle{splncs04}
\bibliography{Bibliography-File}

\section{Additional Results}

\begin{figure}[htb!]
	\centering
	\includegraphics[width=12cm]{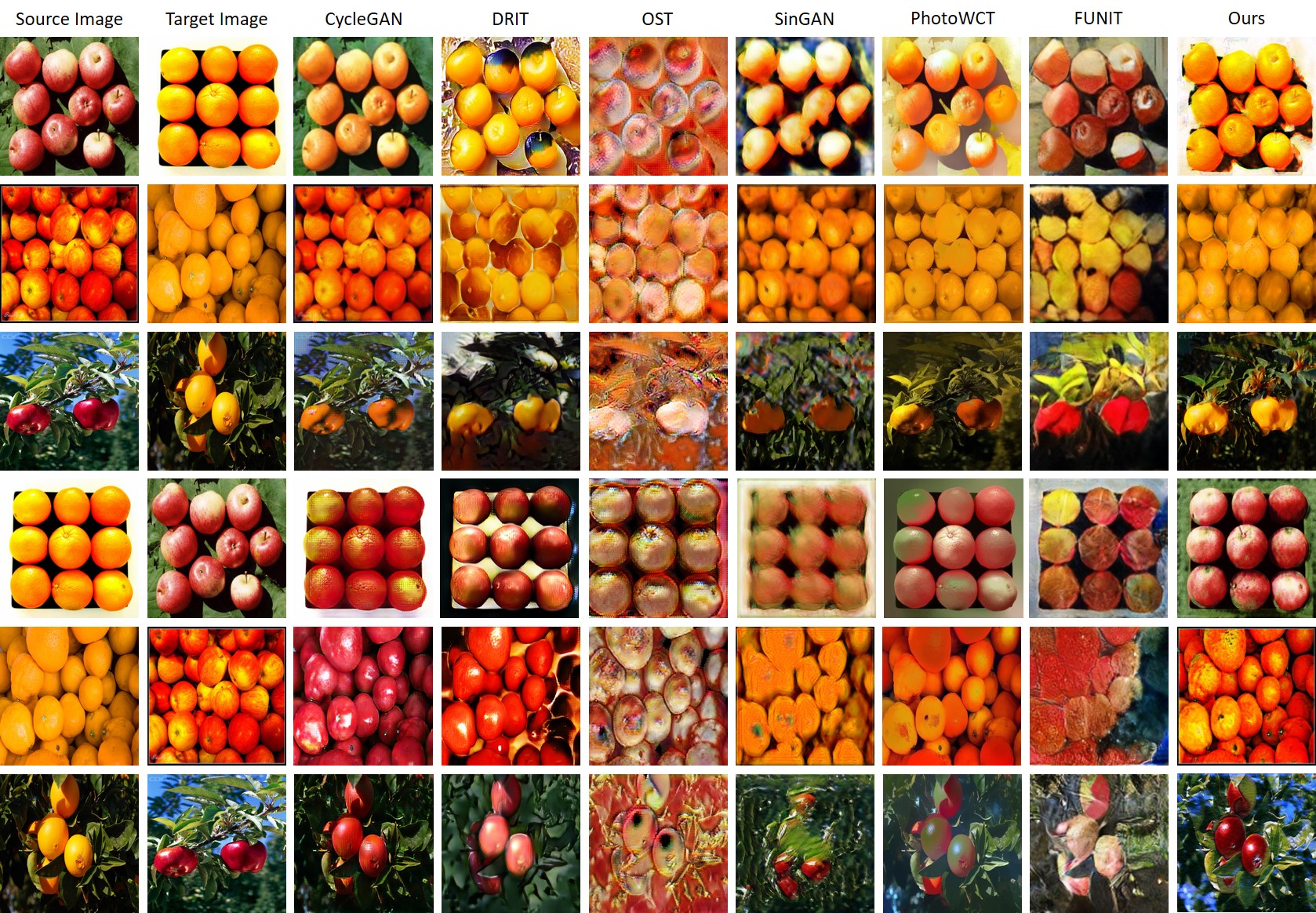}
	\caption{ The top three rows are Apple$\rightarrow$Orange results and the bottom three rows are  Orange$\rightarrow$Apple results. }
	\label{fig:apple2orange}
\end{figure}

\begin{figure}[htb!]
	\centering
	\includegraphics[width=12cm]{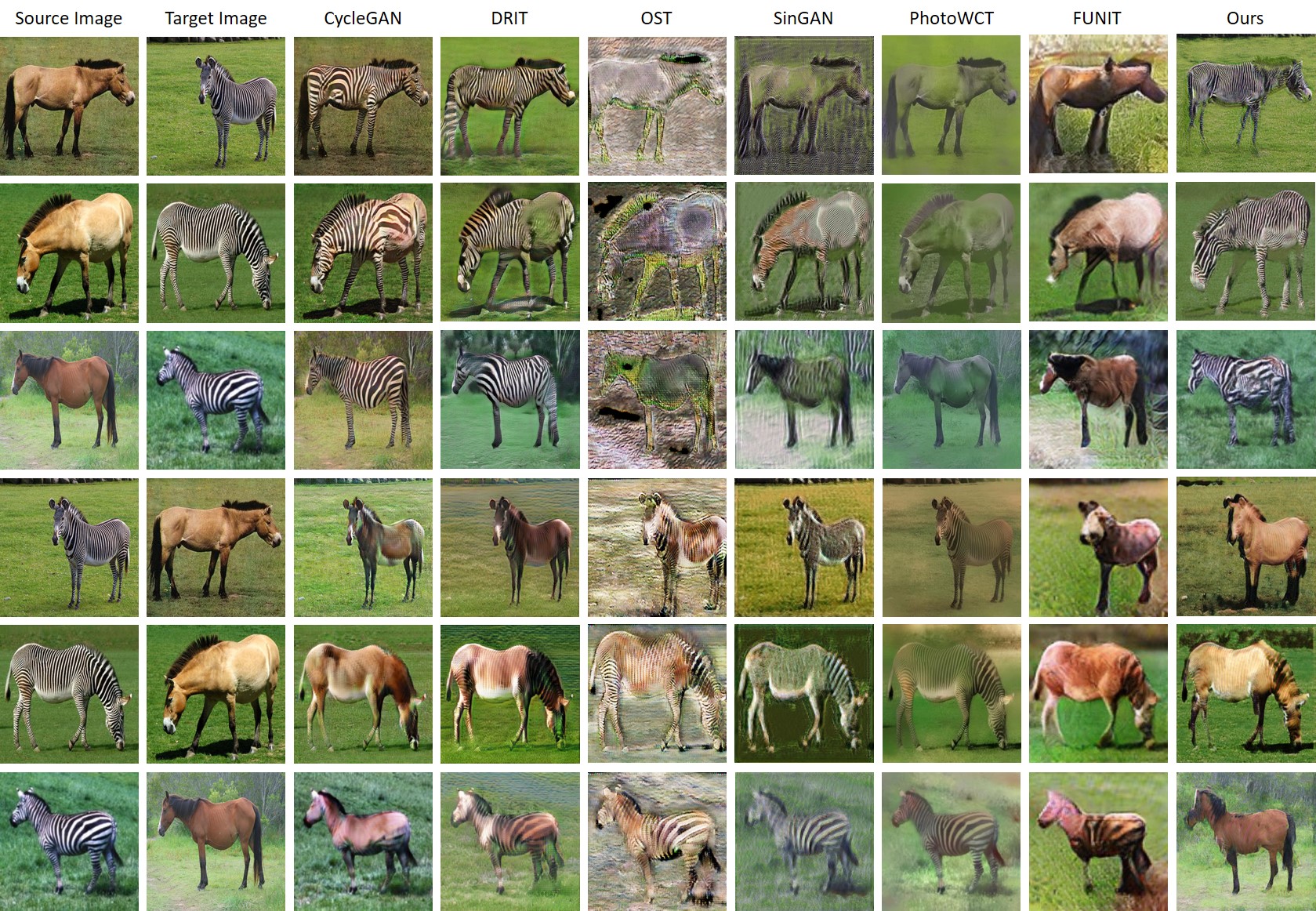}
	\caption{ The top three rows are Horse$\rightarrow$Zebra results and the bottom three rows are  Zebra$\rightarrow$Horse results.}
	\label{fig:horse2zebra}
\end{figure}

\begin{figure}[htb!]
	\centering
	\includegraphics[width=12cm]{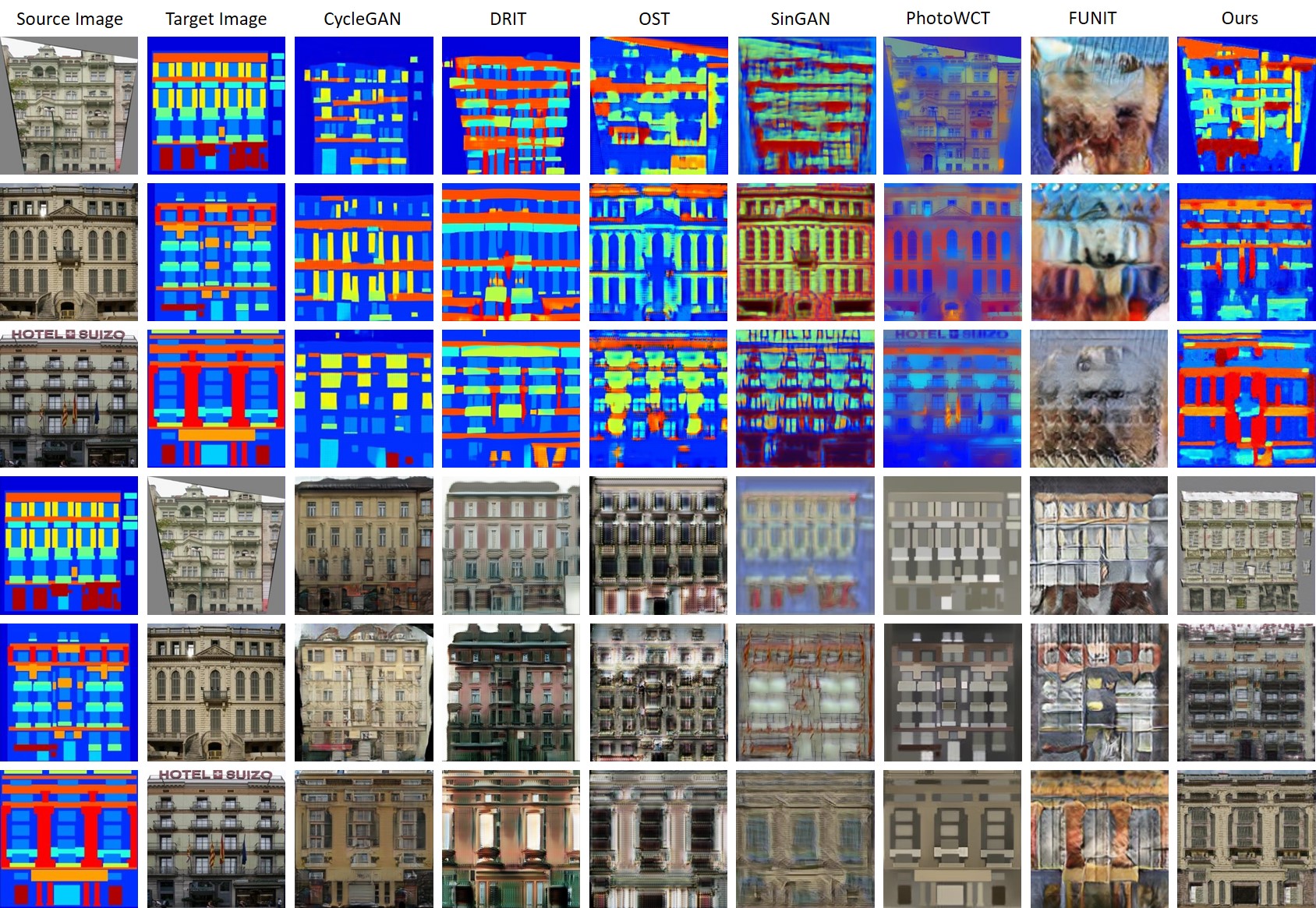}
	\caption{The top three rows are Facade$\rightarrow$Labels results and the bottom three rows are  Labels$\rightarrow$Facade results.}
	\label{fig:facades}
\end{figure}

\begin{figure}[!t]
	\centering
	\includegraphics[width=12cm]{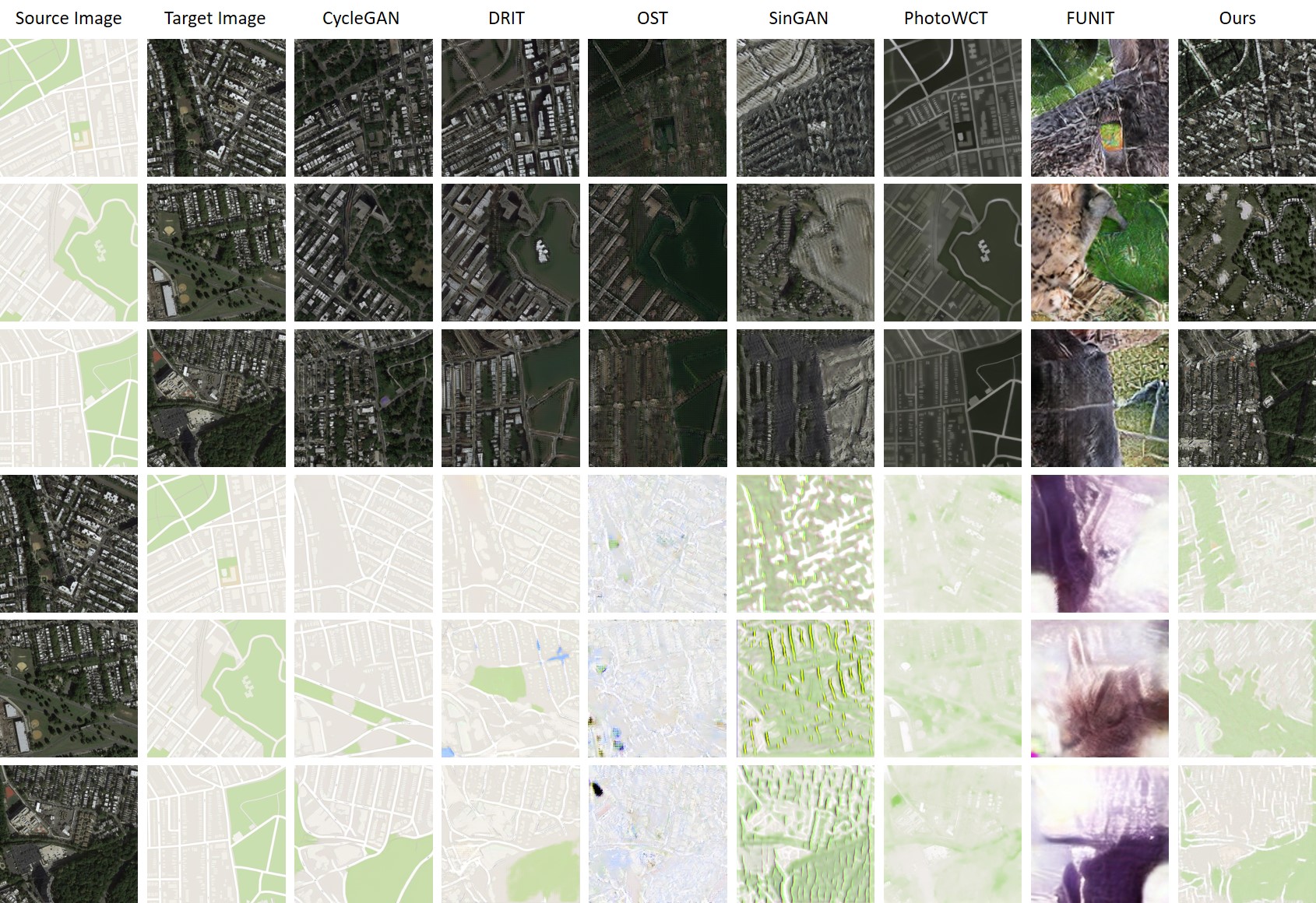}
	\caption{ The top three rows are Map$\rightarrow$Aerial Photo results and the bottom three rows are  Aerial Photo$\rightarrow$Map results.}
	\label{fig:maps}
\end{figure}

\begin{figure}[!t]
	\centering
	\includegraphics[width=12cm]{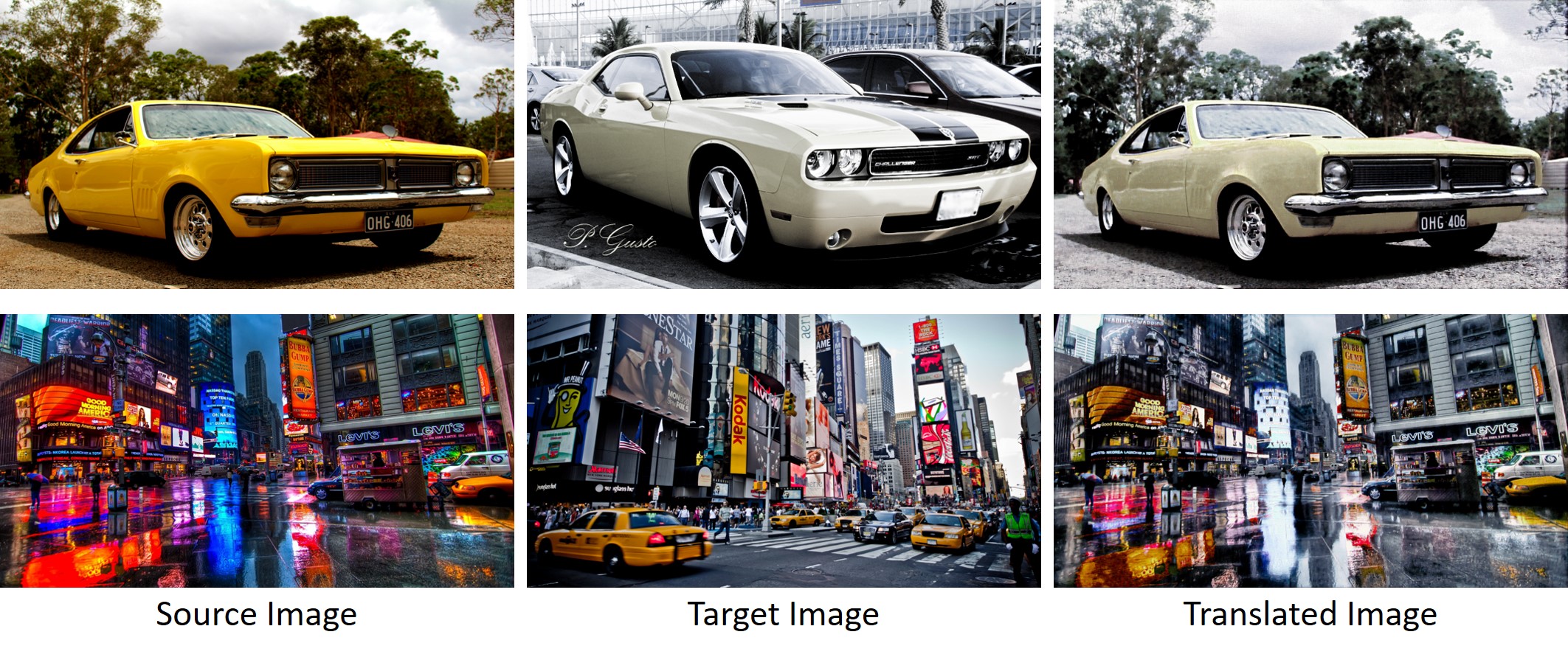}
	\caption{Additional photorealistic style transfer results.}
	\label{fig:photostyle}
\end{figure}

\begin{figure}[!t]
	\centering
	\includegraphics[width=12cm]{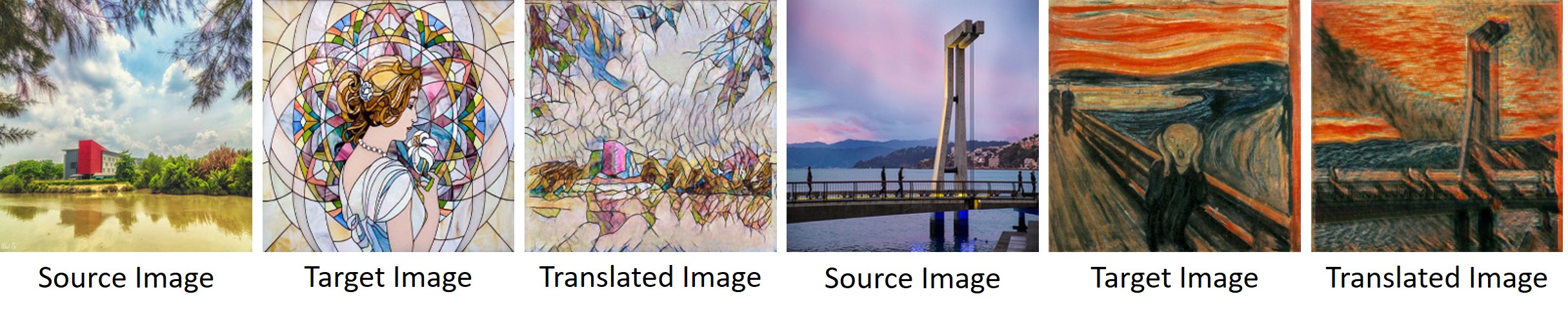}
	\caption{Additional art style transfer results.}
	\label{fig:artstyle}
\end{figure}

\begin{figure}[!t]
	\centering
	\includegraphics[width=12cm]{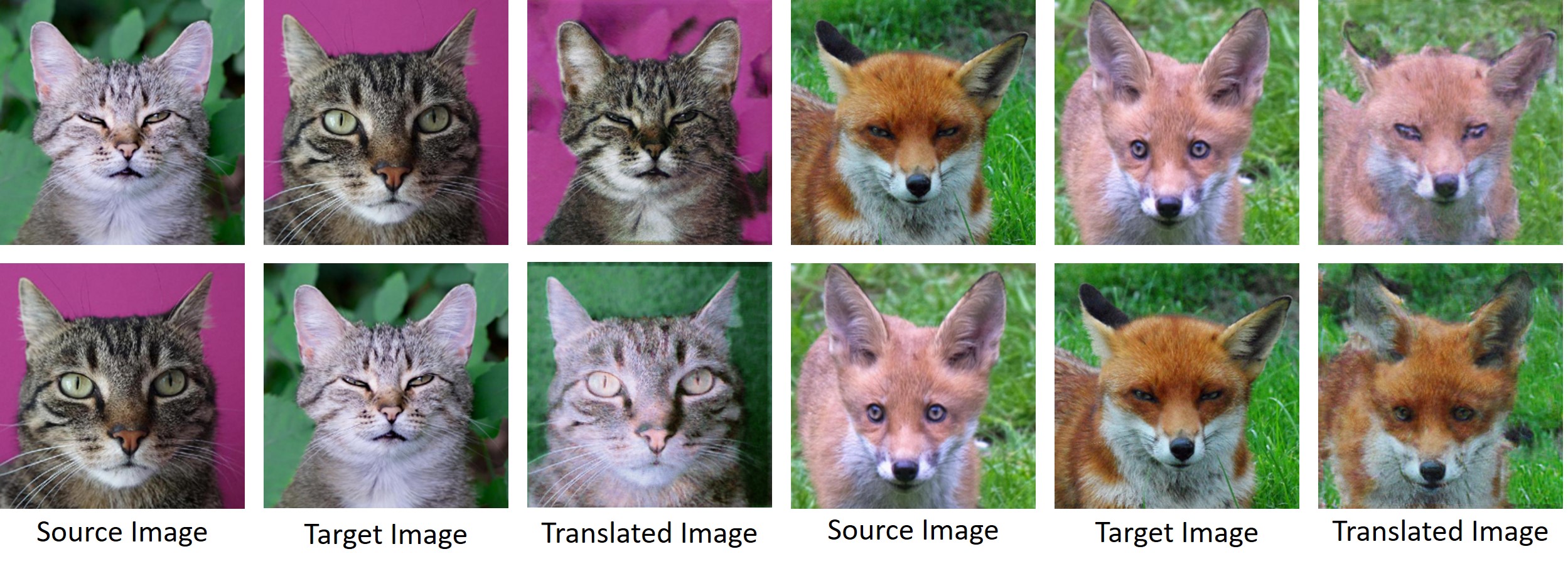}
	\caption{Additional animal face translation results.}
	\label{fig:animalface}
\end{figure}

In this section, we  provide additional results of four general unpaired image-to-image translation tasks: Apple$\leftrightarrow$Orange in Fig. \ref{fig:apple2orange}, Horse$\leftrightarrow$Zebra in Fig. \ref{fig:horse2zebra}, Facade$\leftrightarrow$Labels in Fig. \ref{fig:facades}, and Map$\leftrightarrow$Aerial Photo in Fig. \ref{fig:maps}. From the additional results provided, we can further verify that most of the translation results generated by TuiGAN are better than OST, SinGAN, PhotoWCT and FUNIT. Compared with CycleGAN and DRIT trained
on full data, our model can also achieve comparable performance in many cases. In addition, more results on photorealistic style transfer can be found in Fig. \ref{fig:photostyle}, art style transfer can be found in Fig. \ref{fig:artstyle} and animal face translation can be found in Fig. \ref{fig:animalface}.


\end{document}